\documentclass{article}

\usepackage[preprint,nonatbib]{neurips_2020}

\usepackage[utf8]{inputenc} 
\usepackage[T1]{fontenc}    
\usepackage{hyperref}       
\usepackage{url}            
\usepackage{booktabs}       
\usepackage{amsfonts}       
\usepackage{nicefrac}       
\usepackage{microtype}      
\usepackage{graphicx}
\usepackage{color}
\usepackage{subfigure}
\usepackage{floatrow}
\usepackage{amsmath}

\newcommand{\etal}{\textit{et al.}}
\newcommand{\first}[1]{\textcolor{red}{\textbf{#1}}}
\newcommand{\second}[1]{\textcolor{blue}{\textbf{#1}}}
\newcommand{\third}[1]{\textcolor{green}{\textbf{#1}}}
\newcommand{\latinword}[1]{{\fontfamily{lmtt}\selectfont #1}}

\title{Tracking the Untrackable}

\author{
	Fangyi Zhang \\
	Institue of Computing Technology, CAS, China \\
	\texttt{fangyi.zhang@vipl.ict.ac.cn} \\
}

\begin{document}

\maketitle

\begin{abstract}
Although short-term fully occlusion happens rare in visual object tracking, most trackers will fail under these circumstances. However, humans can still catch up the target by anticipating the trajectory of the target even the target is invisible. Recent psychology also has shown that humans build the mental image of the future. Inspired by that, we present a HAllucinating Features to Track(HAFT) model that enables to forecast the visual feature embedding of future frames. The anticipated future frames focus on the movement of the target while hallucinating the occluded part of the target. Jointly tracking on the hallucinated features and the real features improves the robustness of the tracker even when the target is highly occluded. Through extensive experimental evaluations, we achieve promising results on multiple datasets: OTB100, VOT2018, LaSOT, TrackingNet, and UAV123.
\end{abstract}

\section{Introduction}
Visual object tracking is pursuing the goal of estimating the position and size of arbitrary object in a video sequence. Given the first frame of the unknown target, we need to track the object in subsequent frames. Most recent approaches tackle the tracking problem solely relying on appearance features\cite{PAMI15KCF, ECCV16SiamFC, CVPR18SiamRPN}. However, visual features alone fails when the target under severe occlusion and easily draft to distractor or stay on the previous estimated position.

Lack of motion information cause the tracker fails when severe occlusion happens has been noticed by recent research\cite{icpr16deepmotion,cvpr18flowtrack}. These methods add motion information to the tracker, such as optical flow. They first extract optical flow either form the time-consuming TV-L1 algorithm\cite{perez2013tv} or FlowNet\cite{cvpr15flownet}. With the flow field, motion information can be easily integrated into the tracking model. Tracking is then performed by combing the motion cues and appearance cues. 

However, optical flow need the visual appearance to be consistent\cite{perez2013tv}. When the target under severe deformation or occlusion, the estimated flow field contains much noise, the region of flow field corresponding to the object cannot reflect rich information of the movement of the object. In addition, optical flow calculates the similarity in a given grid which further limits its usage when object have large movements. Another approach\cite{wang2019prediction} incorporate motion information from a different view. It uses Kalman Filter to estimate the target trajectory. The key limitation of this method is that it cannot learn object motions from large video datasets, which restricts its further improvement.

Our goal is to learning the object motion from large scale video dataset while hallucinating the occluded part of the target. Recent psychological theories have shown that human visual system is predictive in nature\cite{greve2015role,summerfield2014expectation}. Greve\cite{greve2015role} has shown that humans build the mental image of the future before initiating muscle movements or motor controls. These representations capture both visual and temporal information of the expected future. Olson \etal\cite{olson2004neuronal} found human will maintain a representation of a target even the moving target was not visible, having behind an occluder. Further, Ekman \etal\cite{summerfield2014expectation} designed moving dot sequence experiments proving that perception is guided by the anticipation of future events. 

Mimicking this biological process, we propose a HAllucinating Features to Track method named HAFT to solve the occlusion problems in visual object tracking. By predicting the future movement of target and hallucinate the occluded part, our method learns to anticipate target representations even under severe occlusion. Specially, we utilize Generative Adversarial Networks(GANs)\cite{nips14gan} to anticipated the movement and representation of target. On the one hand, GANs can make the predicted features more realistic. On the other hand, it can also learn to hallucinate the occluded part of the target. The generator in our GAN framework is GRU\cite{gru} which used to predict the future frame embedding. In order to learn the spatial-temporal features, we incorporate ConvGRU\cite{ICLR15ConvGRU}. As noted, we forecast the future frame in feature level instead of directly predicted the future frame in pixel level. In this way, we can avoid redundant computation which first generates future frame then extracts features. The final feature used to represent target at current time is the fusing of the hallucinated feature and the real feature. Our model achieves promising results on VOT2018\cite{vot18}, OTB100\cite{OTB100}, LaSOT\cite{lasot}, TrackingNet\cite{trackingnet}, and UAV123\cite{uav}.

\begin{figure}[t]
\begin{center}
\includegraphics[width=0.9\linewidth]{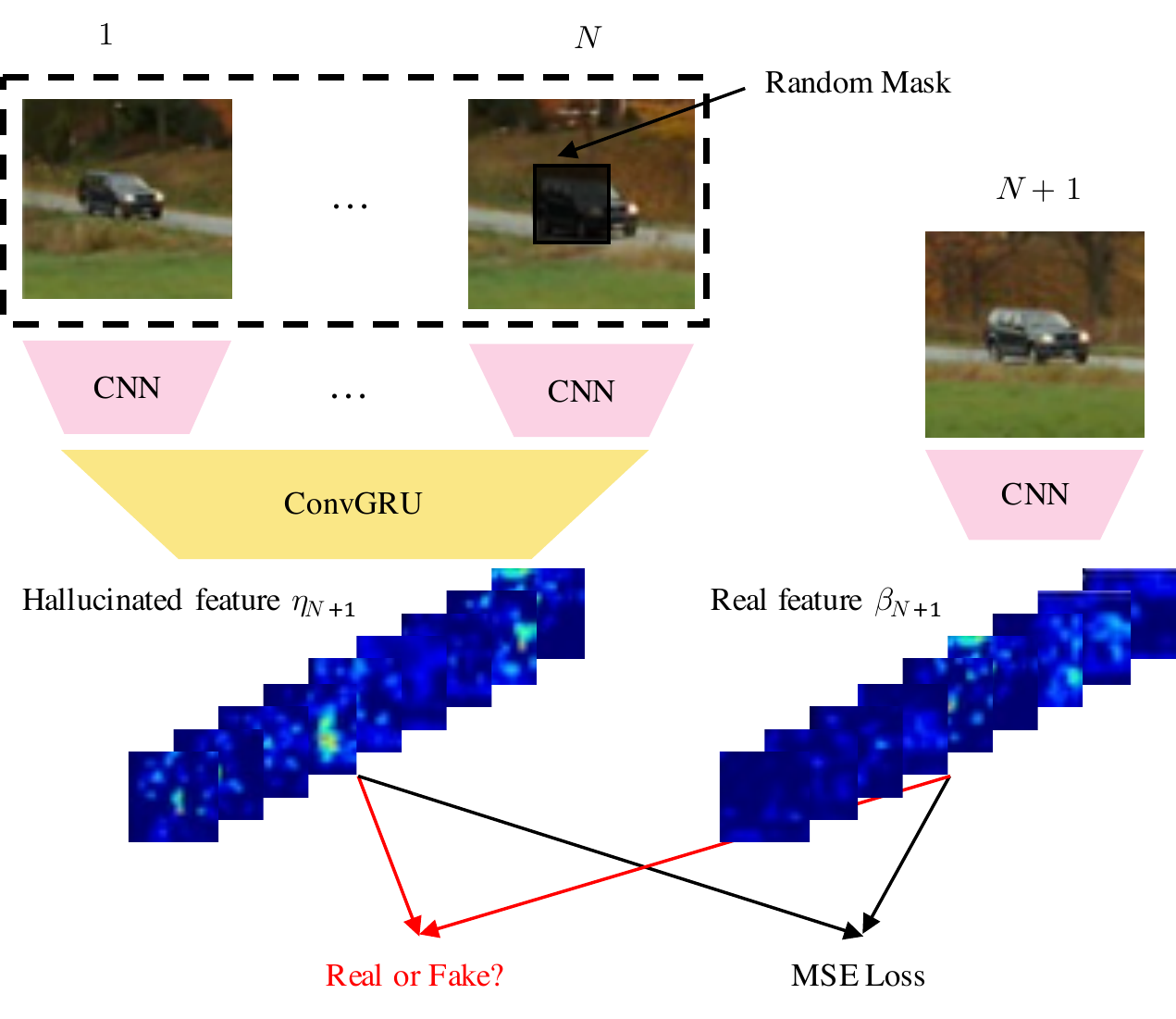}
\end{center}
\vspace{-4mm}
\caption{The structure of HAFT. Here we omit target localization and size estimation part. Our model uses GAN loss and MSE loss to make the hallucinated feature $\eta_{N+1}$ similar to the real feature $\beta_{N+1}$. Due to the fact that occlusion is rare and hard to label, we use random mask to simulate the target being occluded.}
\label{fig:structure}

\end{figure}

\section{Approach}

Our future frame anticipation model is designed to predict future frame while online tracking. The model aims to generate embedding for future frames and obtain a complete target even when the target is fully occluded. Fig. \ref{fig:structure} illustrates the architecture of our model. The proposed network consists of fully convolutional layers (\latinword{conv1-4}) for constructing feature map. Given a video sequence with a set of bounding boxes corresponding to each frame, the network computes a feature map of the input image through a single forward. The feature maps are then sent to a ConvGRU\cite{ICLR15ConvGRU} network to learn a temporal representation. Due to the fact that occlusion happens rare in video sequence, for example 1.86\% frames are marked as heavy occlusion in GOT-10k\cite{got10k} dataset. When training our model, we add random mask to the target to simulate the target being occluded. Generative Adversarial loss and $L_2$ loss are utilized to make the hallucinated features more realistic.

\subsection{HAllucinating Features to Track(HAFT)}

\textbf{Temporal prediction.}
The Recurrent Neural Networks(RNNs) are used to do temporal prediction. However RNNs cannot capture long-term temporal dependency. GRU\cite{gru} is further proposed to solve the above issue. The traditional GRU can only handle flatten features, which not suitable for video representation learning. ConvGRU replace the fully connected layer with convolutional layer, which makes it more efficient for learning spatial dependency. We use ConvGRU to predict the future frame embedding. 

The input of our model is RGB frames, which does not contain any additional motion information. If the input video clip has $N$ frames, the entire input can be represent as:
\begin{equation}
	I = \{ I_1, I_2, \cdots, I_N\}	
\end{equation}
We use feature extractor $\phi$ to frame-wisely extract feature $\alpha$:
\begin{align}
	\alpha & = \{\phi(I_1), \phi(I_2), \cdots, \phi(I_N)\} \\
		   & = \{\alpha_1, \alpha_2, \cdots, \alpha_N\}
\end{align}
The feature $\alpha$ then sent to the ConvGRU model, to capture temporal structure of the target. The outputs of ConvGRU defined as:
\begin{align}
	\hat{\alpha}_t & = \mathrm{CONCAT}[\alpha_1, \alpha_t] \\
	\eta_{t+1} & = \mathrm{ConvGRU}(\hat{\alpha}_t), t = {1, 2, \cdots, T}
\end{align}
where $\eta_{t+1}$ is the anticipated features of next frame. In order to make the predicted features more realistic, we use GANs to train our model.
 
\textbf{Visual and Temporal GANs.}
Recently, GANs are applied in tracking\cite{cvpr2018sint++,cvpr18vital}. These works use GAN to augment positive data. Different from the previous approach, our method use GAN to generate future frame embedding. The original GANs synthesis images without other restriction. On the contrary, our work needs to synthesis future frame embedding which conditioned on the previous frames. The similar work is Conditional-GAN\cite{cgan}, which encode the object label into the generator. As for our algorithm, the ConvGRU model plays the role of the generator in the GAN framework. 

We define the real future video clips $\{I_{2},I_3,\cdots,I_{N+1}\}$, and use the same feature extractor $\phi$ to get the real future features:
\begin{equation}
	\beta = \{\phi(I_2),\phi(I_3),\cdots,\phi(I_{N+1})\}
\end{equation}
This real features $\beta$ are used for GAN training to confuse the discriminator. Our generator $G$ does not directly predict the future frame in pixel level, however, in feature level to speed up the tracking progress. The loss function to train our GAN model is defined as below:
\begin{equation}
	 \ell^V(G^V, D^V) = \sum_{t=1}^{T-1}\log D^V(F_t, \beta^V) + \sum_{t=1}^{T-1}\log(1-D^V(F_t,G^V(F_t)))
\end{equation}
Due the the fact that our discriminator only need to judge the predicted feature is real or fake, we only use three convolution layers, each convolution layer is followed by BatchNorm and LeakyReLU.

In order to stable the GAN training, we use $L_2$ regularization to compare the anticipated features and real features:
\begin{equation}
	    \ell^R = \sum_{t=1}^{T-1} d(h_{t}^V, \beta_{t+1}^V)
\end{equation}

\textbf{Position Localization.}
For target position localization, we need less samples to build a robust model. DiMP\cite{iccv19dimp} utilizes background information, discriminative loss function and high efficient optimizator to achieve this goal. The loss function for localization is defined as below:
\begin{equation}
\ell^L = \sum_{t=2}^{N+1} || r(x_t * f, z_c||^2	
\end{equation}
where $f$ is the filter, $z_c$ gaussian label function which the peak value is at target center, * is the convolution operation. $r$ is the loss function, detailed definition can be found in \cite{iccv19dimp}. $x_t$ is the input features:
\begin{equation}\label{eq:fuse_feat}
	x_t = \lambda \eta_t + (1-\lambda)\beta_t
\end{equation}
where $\lambda$ is used for balance the real feature and the predicted features.

\textbf{Size Estimation.}
The size estimation module used in our model is same as ATOM\cite{CVPR19ATOM} which incorporated the IoU Net\cite{jiang2018acquisition}. The IoU score is estimated by the dot product between template feature and search region feature which extracted from the PrRoI Pooling, then followed by a fully connected layer:
\begin{equation}
	\ell^S = \sum_{t=2}^{N+1} g(\beta_1 \cdot x_t)
\end{equation}
where $\beta_1$ is the template frame feature, $x_t$ is the fused feature defined in Eq. \ref{eq:fuse_feat}, and $g$ is the fully connected layer. The size can be estimated with back propagation to the PrRoI Pooling layer.

\begin{figure}[t]
\begin{center}
\includegraphics[width=0.9\linewidth]{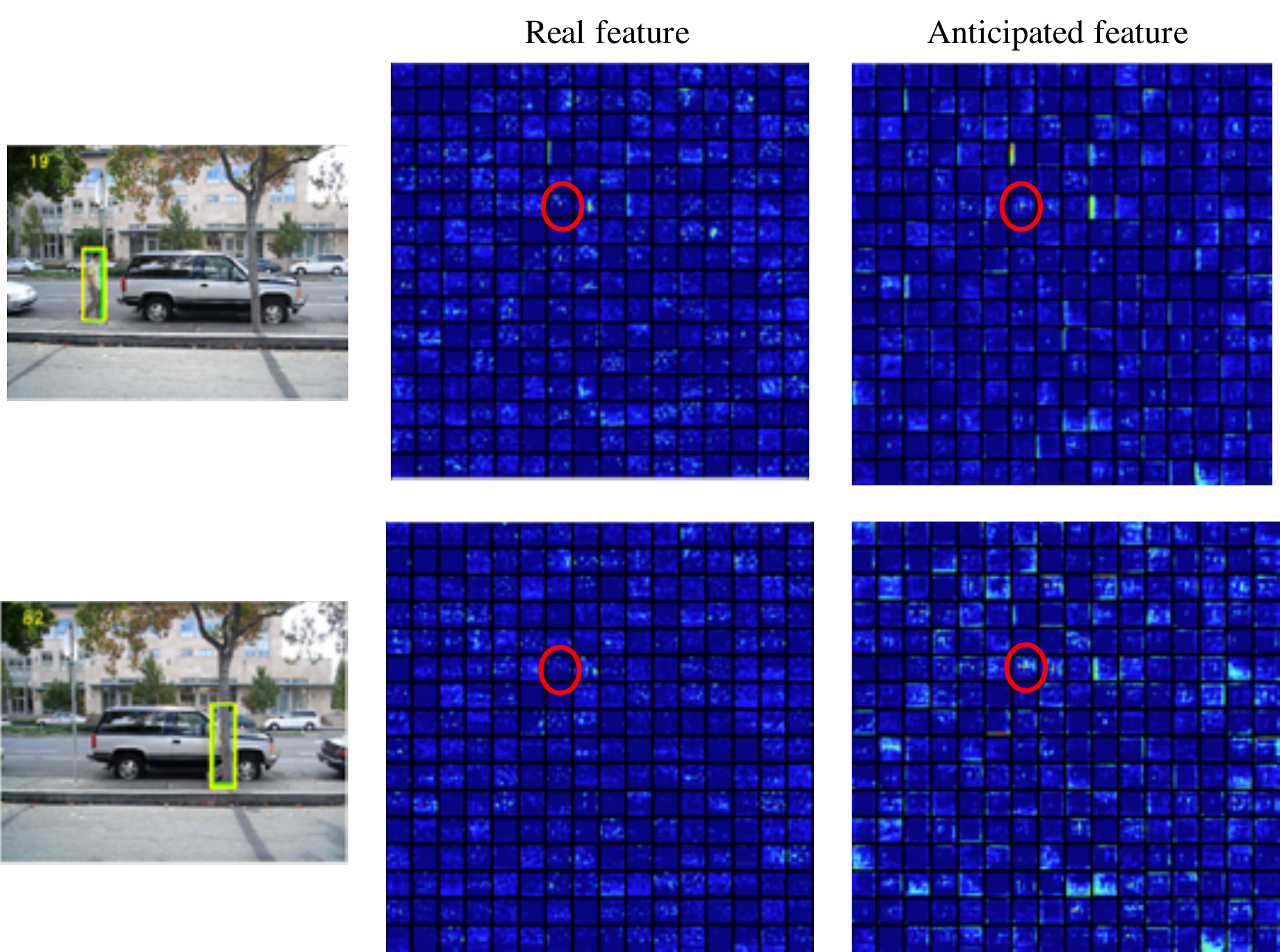}
\end{center}
\vspace{-3mm}
\caption{Visualization of the predicted features. When the target is not occluded in the first row, the response marked by red circle both visible in the real features and predicted features. However, in the second row, when the target is occluded by tree, the response in the real feature is not visible while in the predicted features can be observed.}
\label{fig:vis_occ}
\end{figure}

The final loss function can be written as:
\begin{equation}
	\ell = w^V\ell^V + w^R\ell^R+w^L\ell^L+w^S\ell^S
\end{equation}
where $w^V, w^R, w^L$, and $w^S$ is hyper-parameters to control the each loss's contribution to the final loss. Fig. \ref{fig:vis_occ} is the anticipated features of our model. Even the target is occluded, our model can still predicted the position of the target.

\section{Experiments}

\subsection{Implementation Details}
\textbf{Offline Training.}  Our training procedure is same as DiMP\cite{iccv19dimp}. First, we random sample consequent video clips, the video clips contain 30 frames which then cropped according to the ground-truth bounding boxes. In order to making the training progress close to the tracking progress, the bounding box of the previous frame is used to crop the current frame. Due to the fact that the bounding boxes predicted by the tracker is not accurate, we add random scale and small displacement to the bounding boxes then to create the search region. We also utilize random mask which added to the target to simulate the occlusion.

We initialize the backbone network with DiMP, the feature extractor is ResNet18\cite{he2016deep}. We use training set of TrackingNet\cite{trackingnet}, LaSOT\cite{lasot}, and GOT-10k\cite{got10k} to train our tracker. We sample 20,000 videos one epoch, total training epoch is 50 with ADAM optimizator. The learning rate decays 0.2 every 15 epochs. Our tracker trained on one Nvidia TITAN 2080TI GPU, total training time is 12 hours.

\textbf{Online Tracking.}  
Given the target in the first frames, we construct 15 samples with data augmentation strategy. Then learning a target model as in DiMP. Besides, we initialize ConvGRU model to predicted the features of next frame. In the subsequent tracking progress, we fuse the real feature and anticipated features with Eq. \ref{eq:fuse_feat}, then localize the target and estimate the object size\cite{CVPR19ATOM}.

\subsection{Ablation Analysis}

\begin{figure}[t]\CenterFloatBoxes
\begin{floatrow}
\ffigbox[\FBwidth]
{\caption{OTB100 AUC v.s. $\lambda$.}\label{fig:lambda}}
{\includegraphics[width=1.\linewidth]{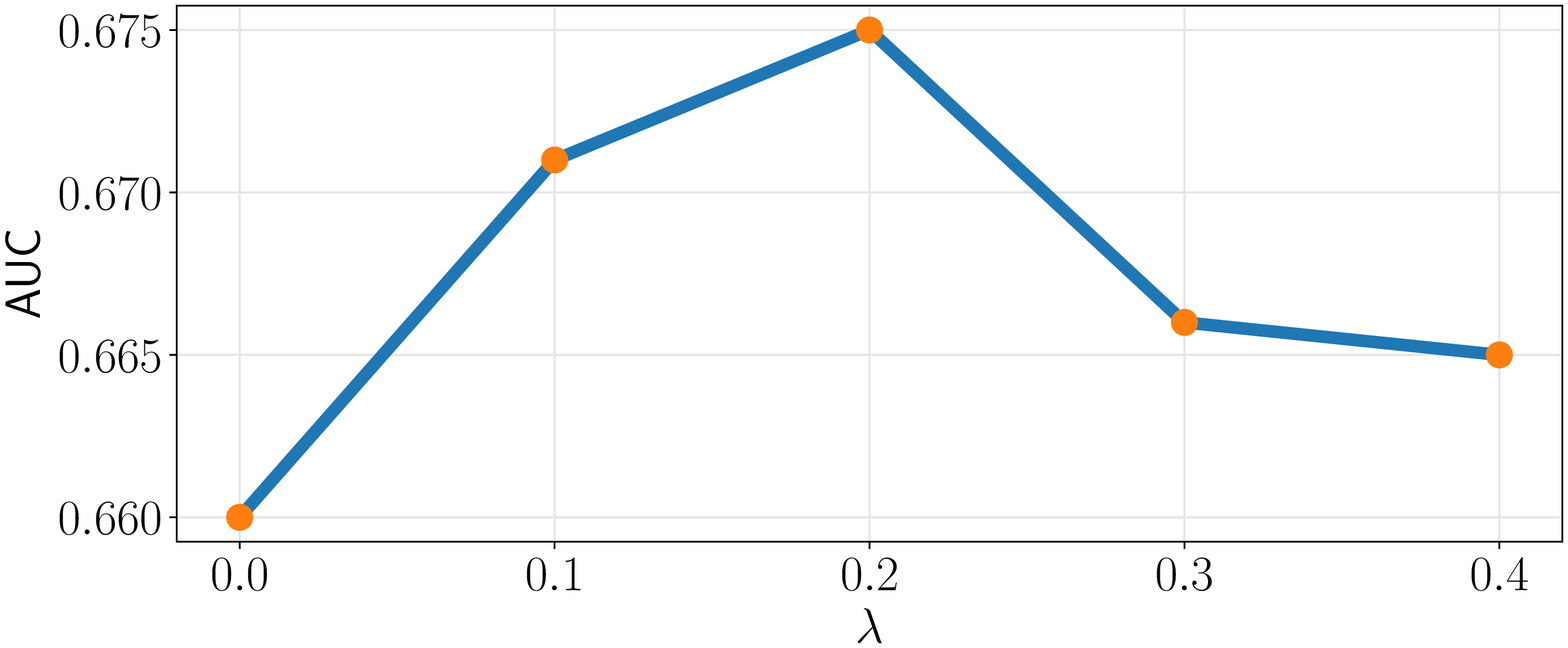}}
\killfloatstyle\ttabbox[\Xhsize]
{\caption{Ablation analysis of OTB100}\label{tab:ablation}}
{
	\setlength{\tabcolsep}{6mm}
	\begin{tabular}{c c | c}
	\toprule
		GAN 		& $L_2$ 		 & AUC   \\
	\midrule
					& 			 & 0.660 \\
		\checkmark  & 			 & 0.551 \\
					& \checkmark & 0.652 \\
		\checkmark  & \checkmark & 0.679 \\
	\bottomrule
	\end{tabular}
}
\end{floatrow}
\end{figure}

\textbf{Different Loss.} In order to verify the effectiveness of the loss, we conduct ablation analysis on OTB100. The results are shown in Fig. \ref{tab:ablation},  Our baseline method is DiMP-18\cite{iccv19dimp}, the AUC score on OTB100 is 0.660. Directly using GAN to generate future frame features will cause the training progress unstable, and leading to severely degenerated tracking performance, the AUC will drop dramatically to 0.551. Also we found directly applying the $L_2$ loss to minimize difference between the predicted features and real future frame features, which will make the features unrealistic, and the tracking performance also drops 1.4\%. Combing both loss, the AUC on OTB100 increased to 0.679. 

\textbf{Fusing Parameter $\lambda$.} We also use OTB100 to get the best $\lambda$ to fuse the predicted features and real features. As shown in Fig. \ref{fig:lambda}, the best $\lambda$ we choose to conduct the subsequent experiments is 0.2.

\subsection{State-of-the-art Comparison}

We compare our method with state-of-the-art methods on several tracking benchmarks, including VOT2018\cite{vot18}, OTB100\cite{OTB100}, LaSOT\cite{lasot}, TrackingNet\cite{trackingnet}, and UAV123\cite{uav}.

\setlength{\tabcolsep}{1mm}
\begin{figure}[t]\CenterFloatBoxes
\begin{floatrow}
\killfloatstyle\ttabbox[\Xhsize]
{\caption{Evaluation results of different trackers on VOT2018. The best top 3 results are marked as \first{red}, \second{blue} and \third{green}.}
 \label{tab:vot18}}
{
	\begin{tabular}{c | c c c c c c c| c}
	\toprule
	Tracker & DRT & RCO & DaSiamRPN	& MFT & ATOM & SiamRPN++ & DiMP-18 & HAFT \\
		& \cite{cvpr18drt} & \cite{vot18} & \cite{eccv2018dasiamrpn} & \cite{vot18} & \cite{CVPR19ATOM} & \cite{CVPR19SiamRPN++} & \cite{iccv19dimp} \\
	\midrule
	Accuracy($\uparrow$)			  &  0.519	&	0.507	& 0.586	 & 0.505		  & \third{0.590} 	& \first{0.600}   & \second{0.594}  & 0.587				\\
	Robustness($\downarrow$)			  &  0.201 	&	0.155	& 0.276	 & \first{0.140}  & 0.204	& 0.234			  & \third{0.182}			& \second{0.155}	\\
	EAO($\uparrow$)					  &  0.356  &	0.376	& 0.383	 & 0.385		  & 0.401	& \second{0.411}  & \third{0.402}			& \first{0.432}		\\
	\bottomrule
	\end{tabular}
}
\end{floatrow}
\end{figure}


\textbf{VOT2018\cite{vot18}.} The VOT2018 dataset consists of 60 challenging videos. Each sequence is per-frame annotated by five visual attributes, and the bounding box is generated from pixel-wise segmentation of the tracked object. In Tab. \ref{tab:vot18} we compare our tracker in terms of Expected Average Overlap(EAO), Accuracy, and Robustness with top-ranked trackers in VOT2018 benchmarks. The proposed tracker achieves the top-ranked performance with respect to EAO. Compared with the baseline tracker DiMP-18, we achieve 7.3\% relative gains on EAO. 

\begin{figure}[t]
\begin{center}
\subfigure{\includegraphics[width=0.49\linewidth]{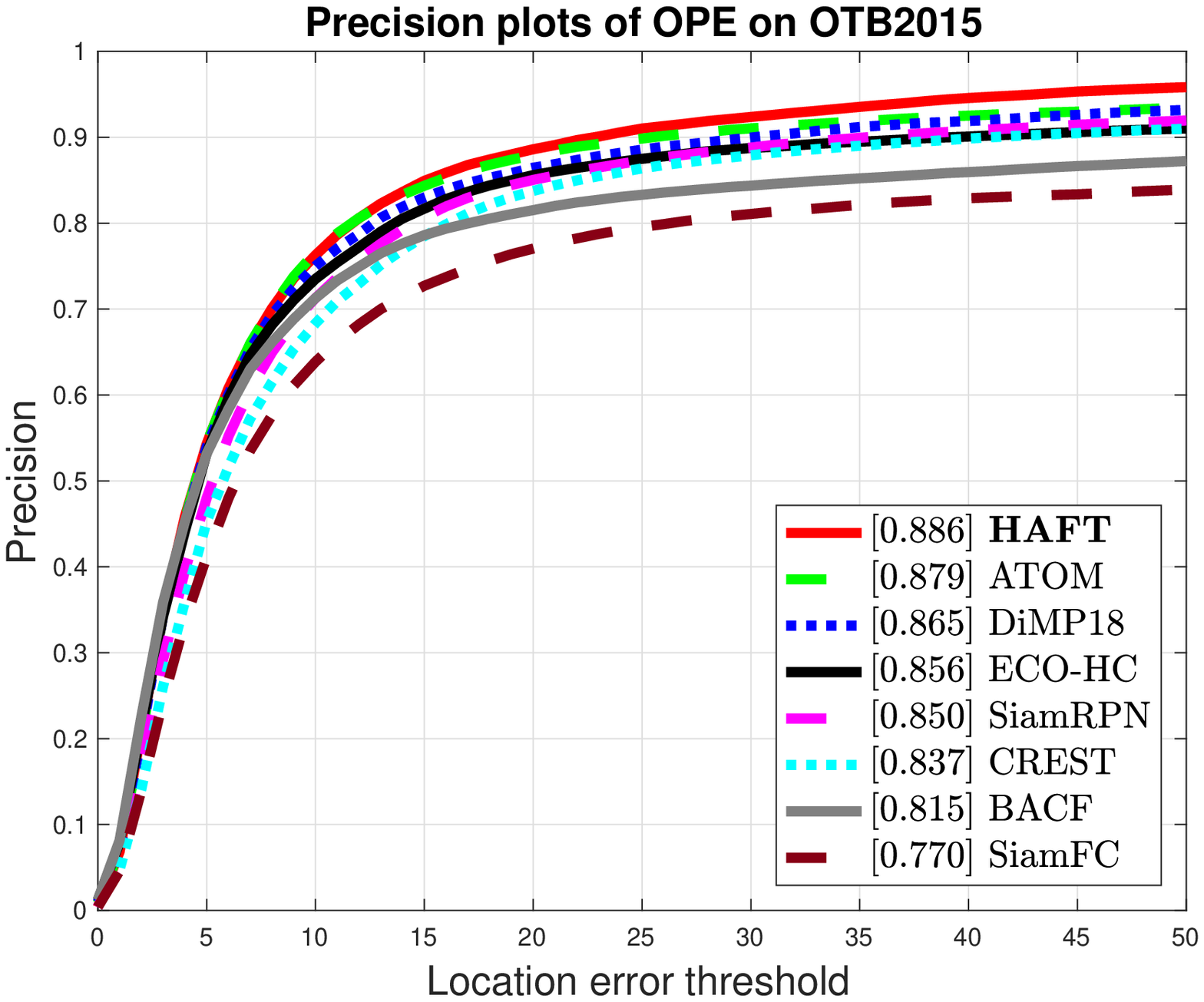}}
\subfigure{\includegraphics[width=0.49\linewidth]{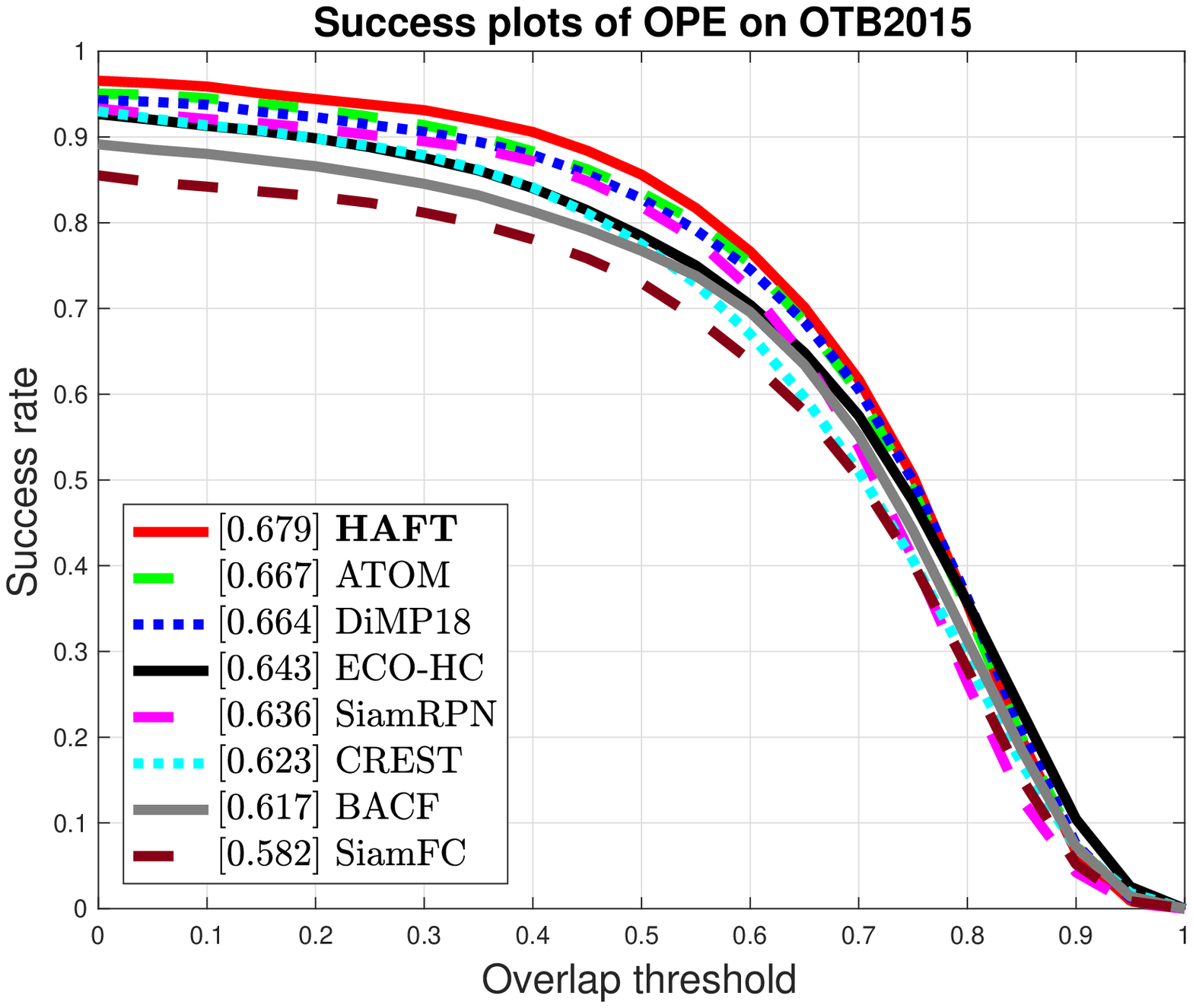}}
\end{center}
\vspace{-1mm}
\caption{Evaluation results of different trackers on OTB100.}
\vspace{-4mm}
\label{fig:otb100}
\end{figure}

\textbf{OTB100\cite{OTB100}.} The OTB100 provides a fair comparison on accuracy and robustness with precision plots and success plots. We compare our trackers with 9 state-of-the-art trackers(ECOHAFTHC\cite{CVPR17ECO}, DaSiamRPN\cite{eccv2018dasiamrpn}, ATOM\cite{CVPR19ATOM}, DiMP-18\cite{iccv19dimp}, SiamRPN\cite{CVPR18SiamRPN}, SiamFC\cite{ECCV16SiamFC}). The precision plots and success plots are shown in Fig. \ref{fig:otb100}

\begin{figure}[t]
\begin{center}
\subfigure{\includegraphics[width=0.43\textwidth]{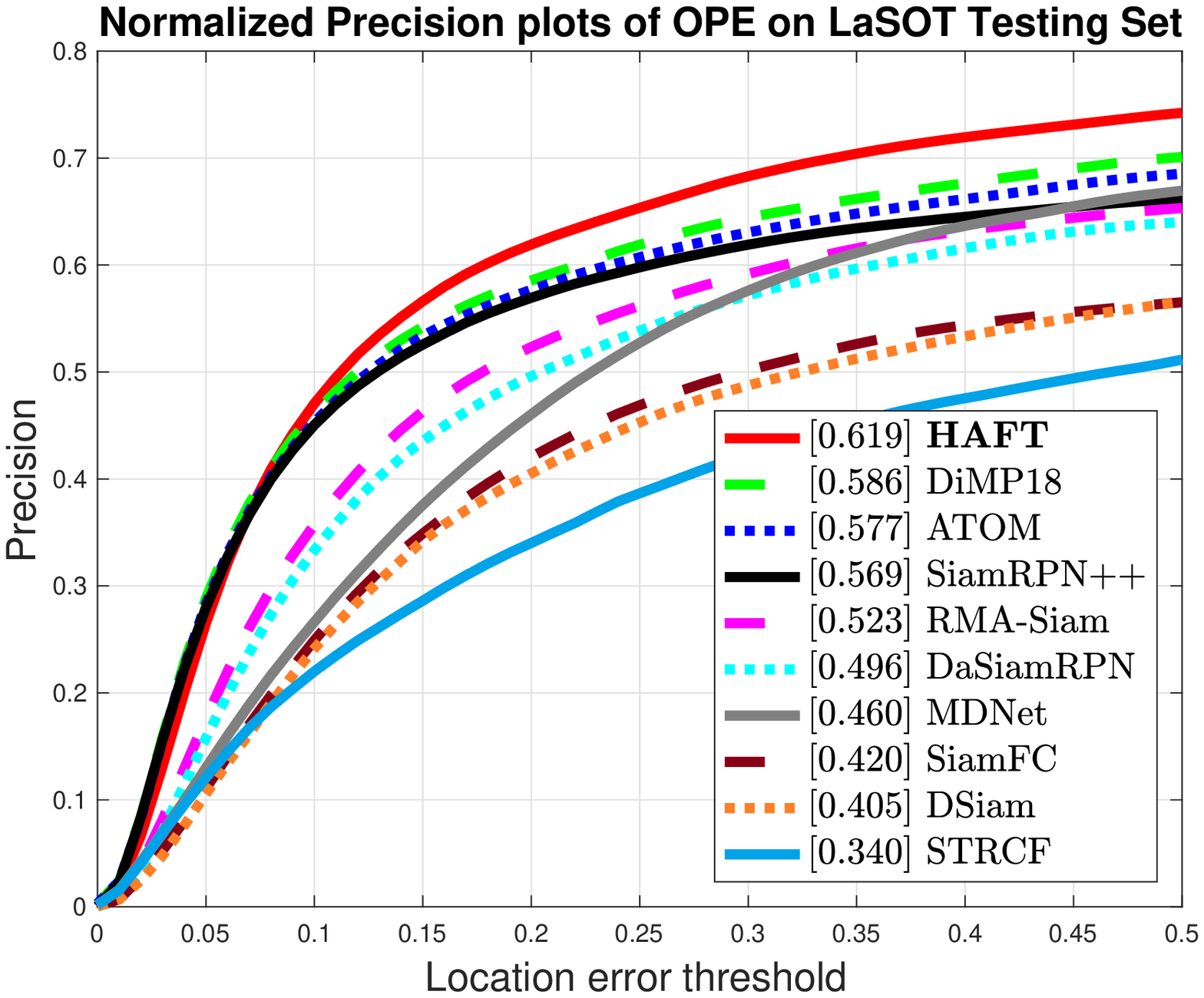}}
\hspace{6mm}
\subfigure{\includegraphics[width=0.43\textwidth]{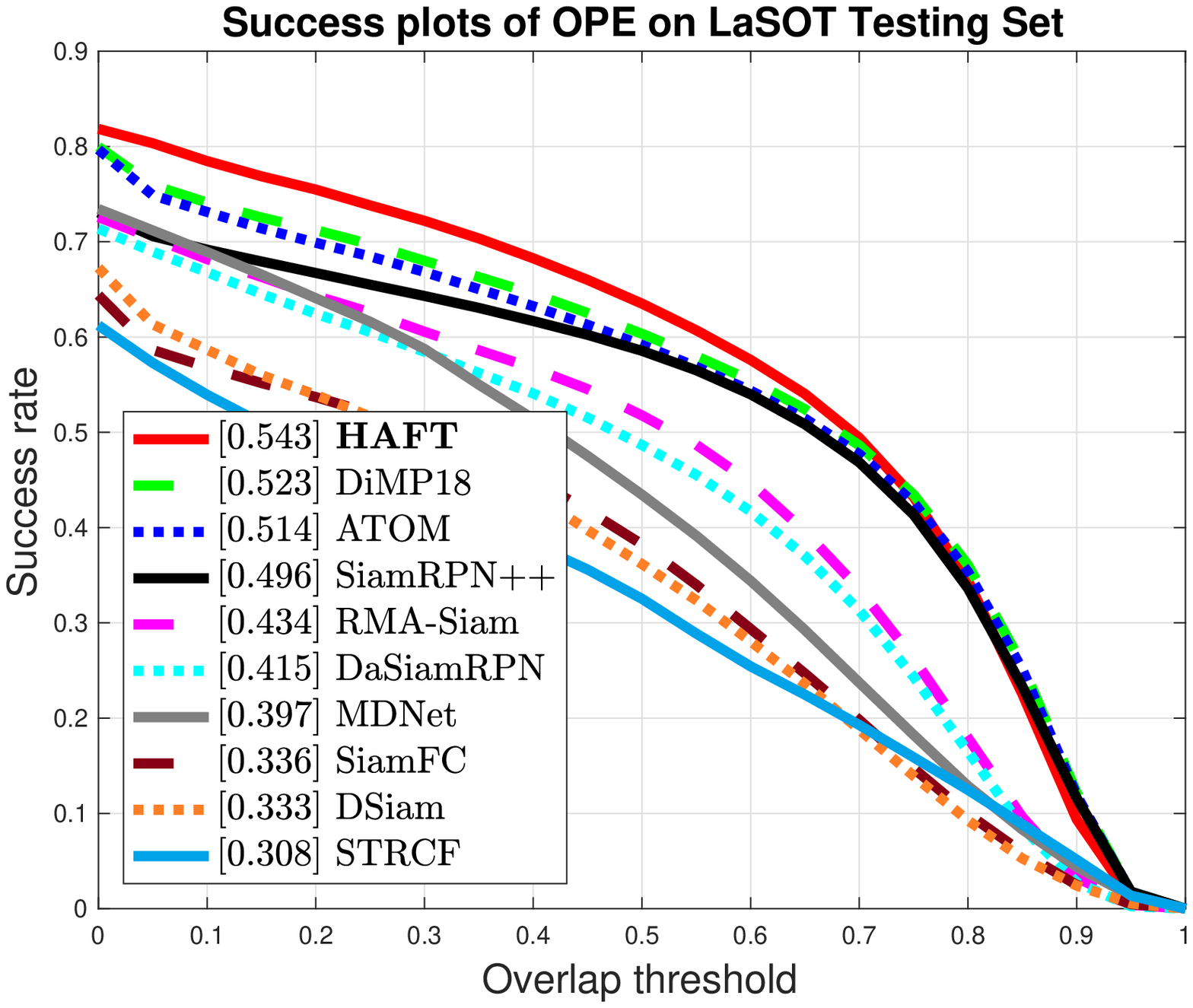}}
\end{center}
\vspace{-4mm}
\caption{Evaluation results of different trackers on LaSOT.}
\label{fig:lasot}
\end{figure}

\textbf{LaSOT\cite{lasot}.} The LaSOT dataset provides a large-scale, high-quality dense annotations with 1400 videos in total. We follow the protocal II which uses 280 testing videos to evaluate our tracker with Normalized Precision Plots and Success Plots. Fig. \ref{fig:lasot} reports the overall performance of our tracker. We compare our tracker with 7 top performance approaches, including MDNet\cite{CVPR16MDNet}, DSiam\cite{iccv17dsiam}, STRCF\cite{cvpr2018strcf}, DaSiamRPN\cite{eccv2018dasiamrpn}, SiamRPN++\cite{CVPR19SiamRPN++}, ATOM\cite{CVPR19ATOM}, and DiMP-18\cite{iccv19dimp}. Our tracker achieves top ranked performance on these three metrics. Compared with the baseline tracker DiMP-18, HAFT achieves 5.6\% relative gains on Normalized Precision Plots and 3.8\% relative gains on Success Plots.

\begin{figure}[t]
\begin{center}

\subfigure{\includegraphics[width=0.23\linewidth]{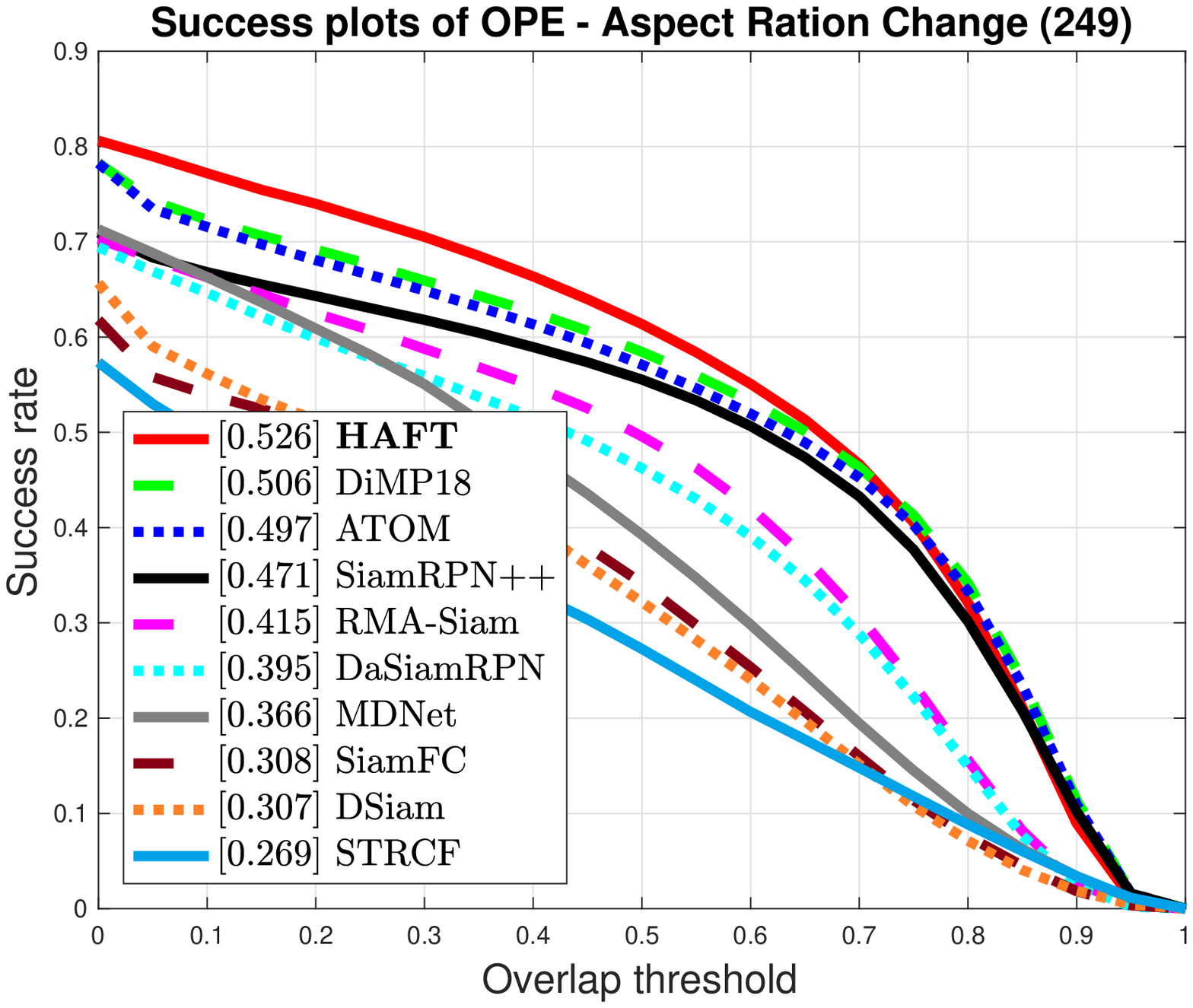}}
\subfigure{\includegraphics[width=0.23\linewidth]{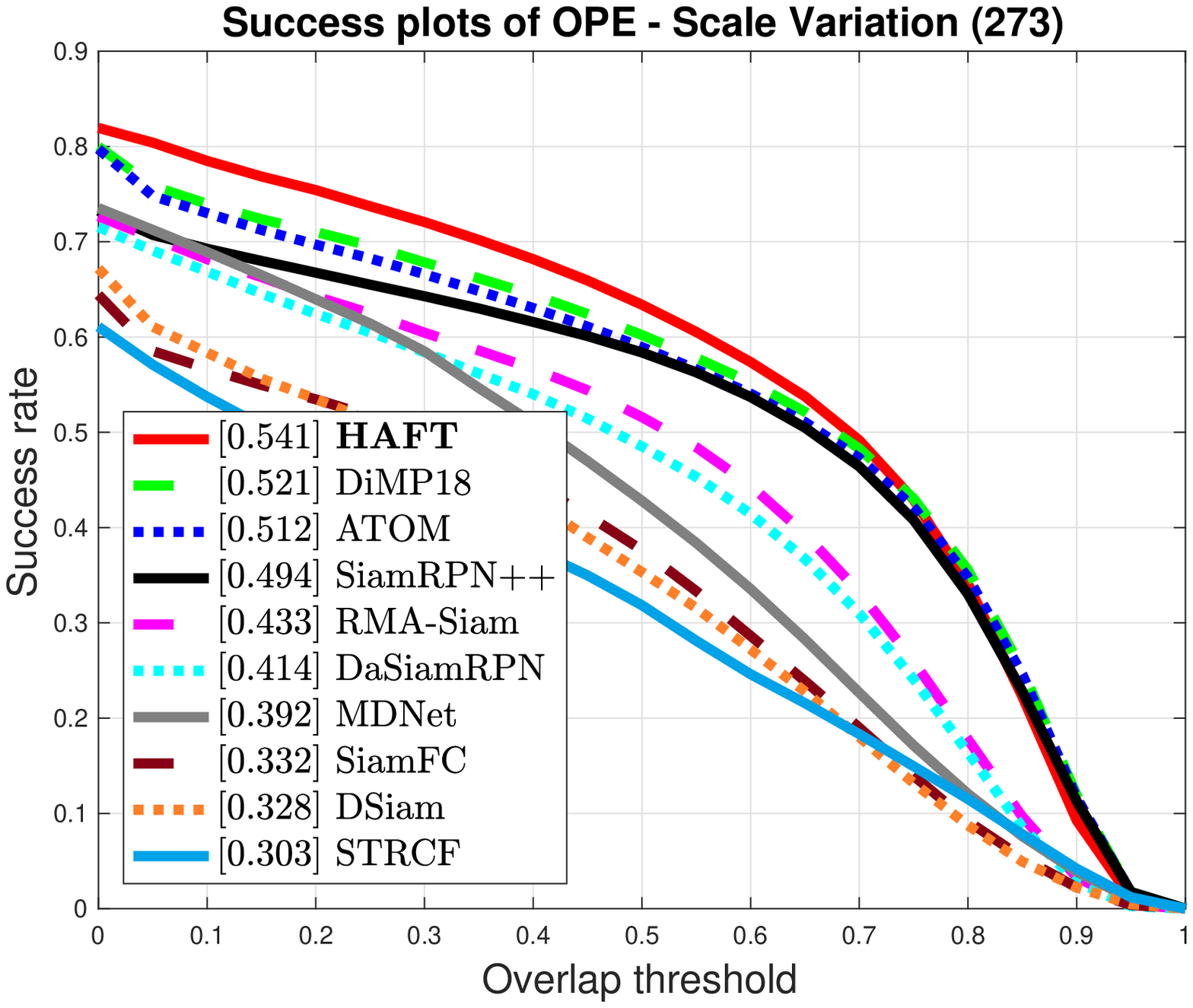}}
\subfigure{\includegraphics[width=0.23\linewidth]{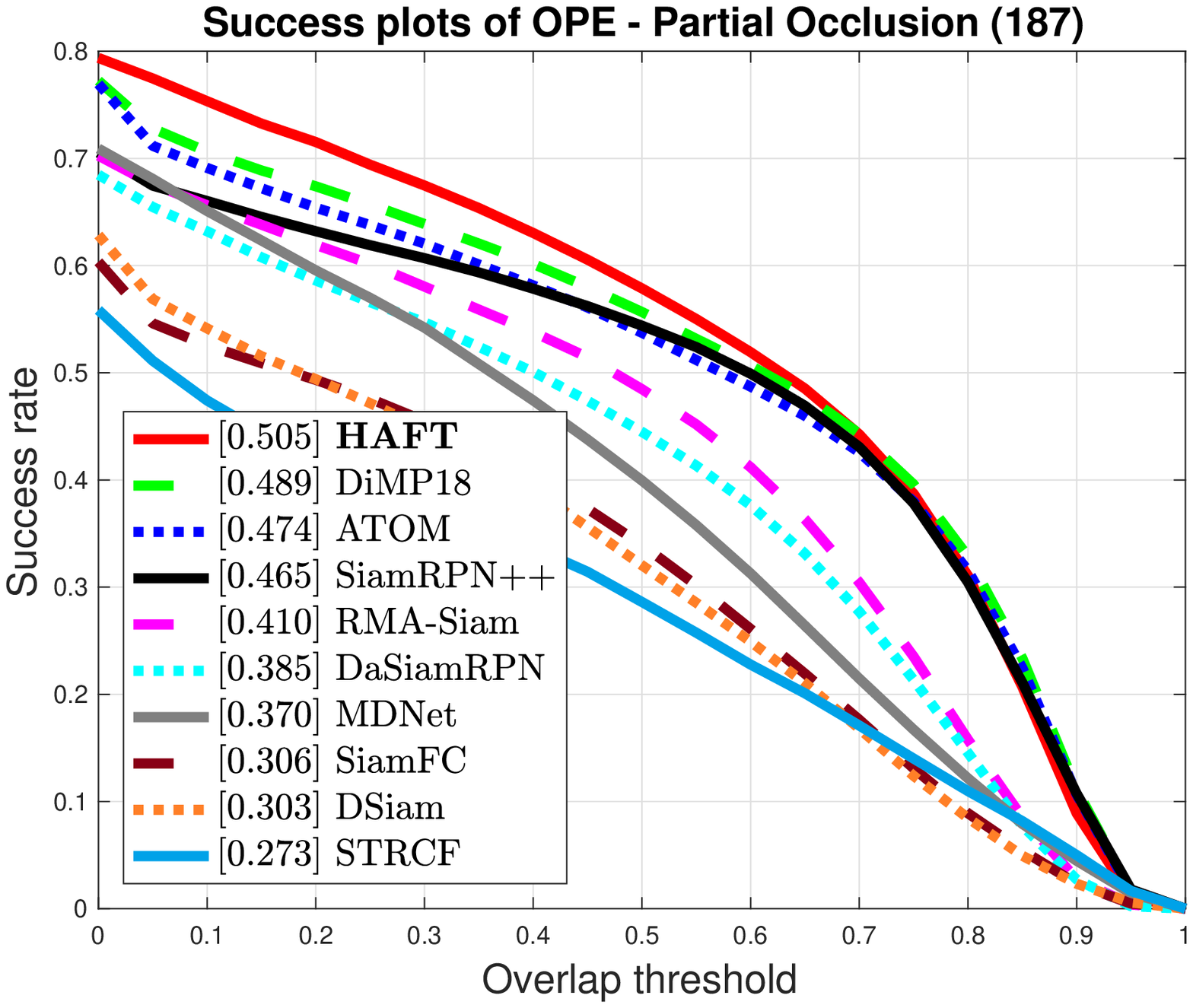}}
\subfigure{\includegraphics[width=0.23\linewidth]{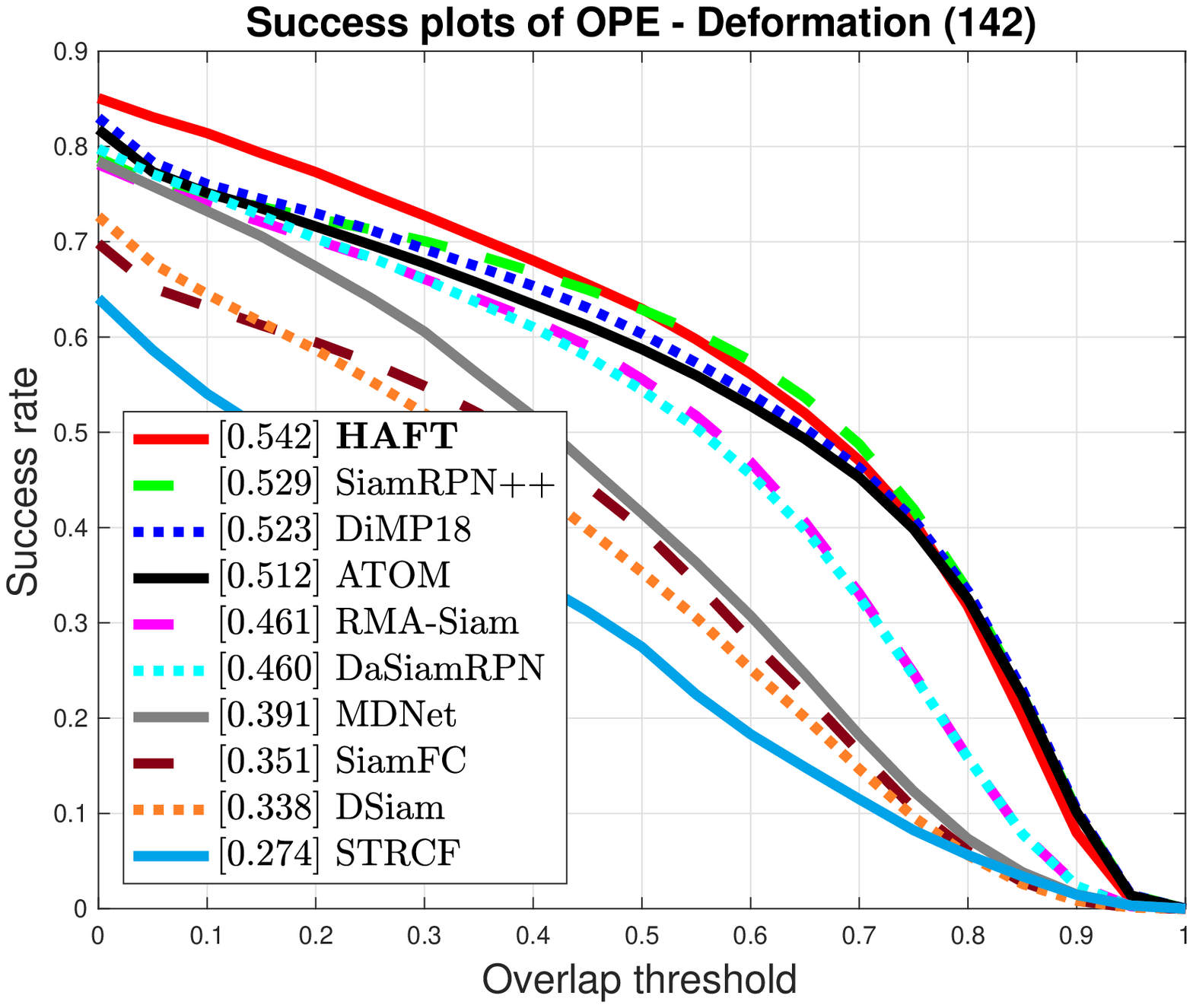}}

\subfigure{\includegraphics[width=0.23\linewidth]{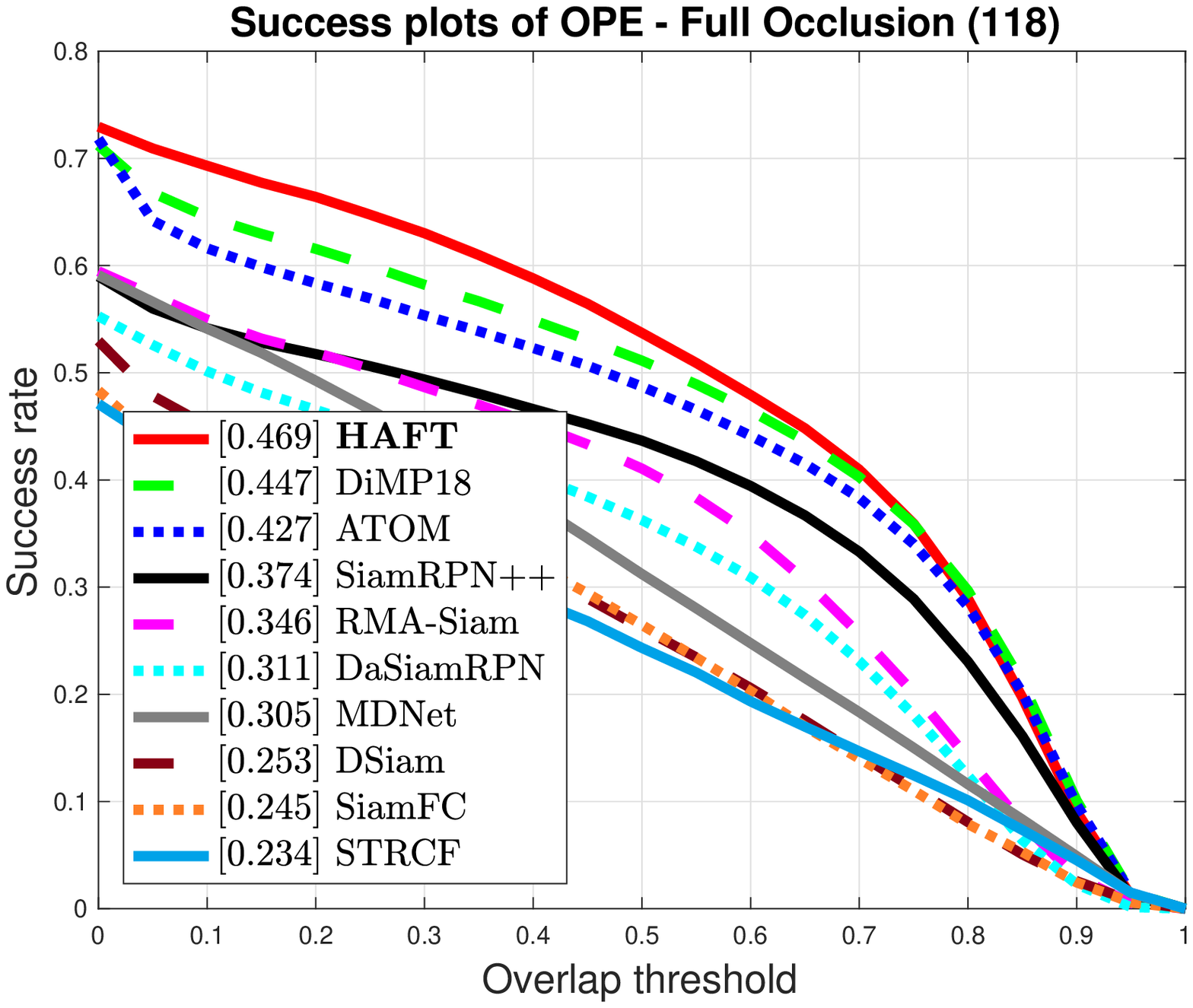}}
\subfigure{\includegraphics[width=0.23\linewidth]{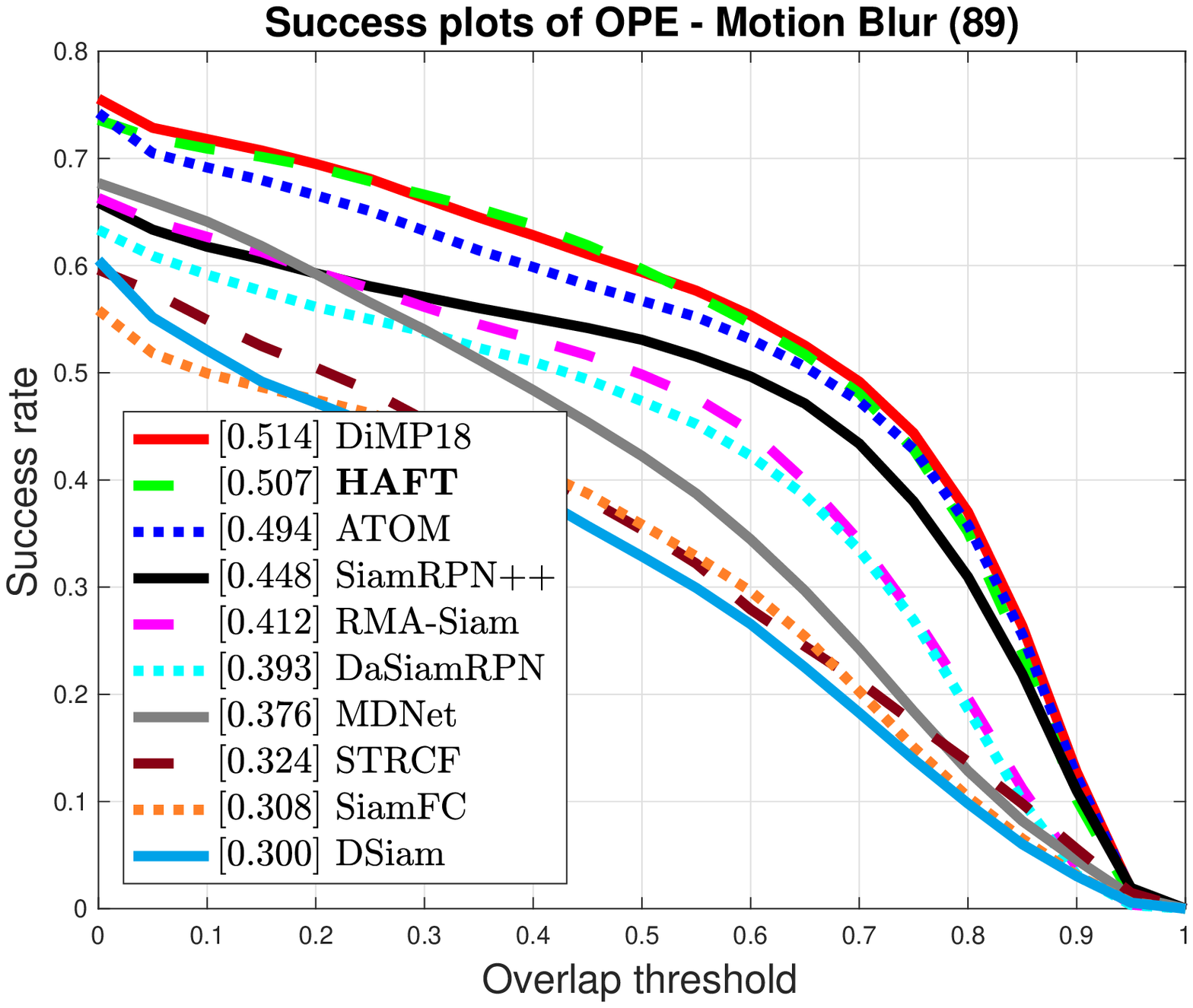}}
\subfigure{\includegraphics[width=0.23\linewidth]{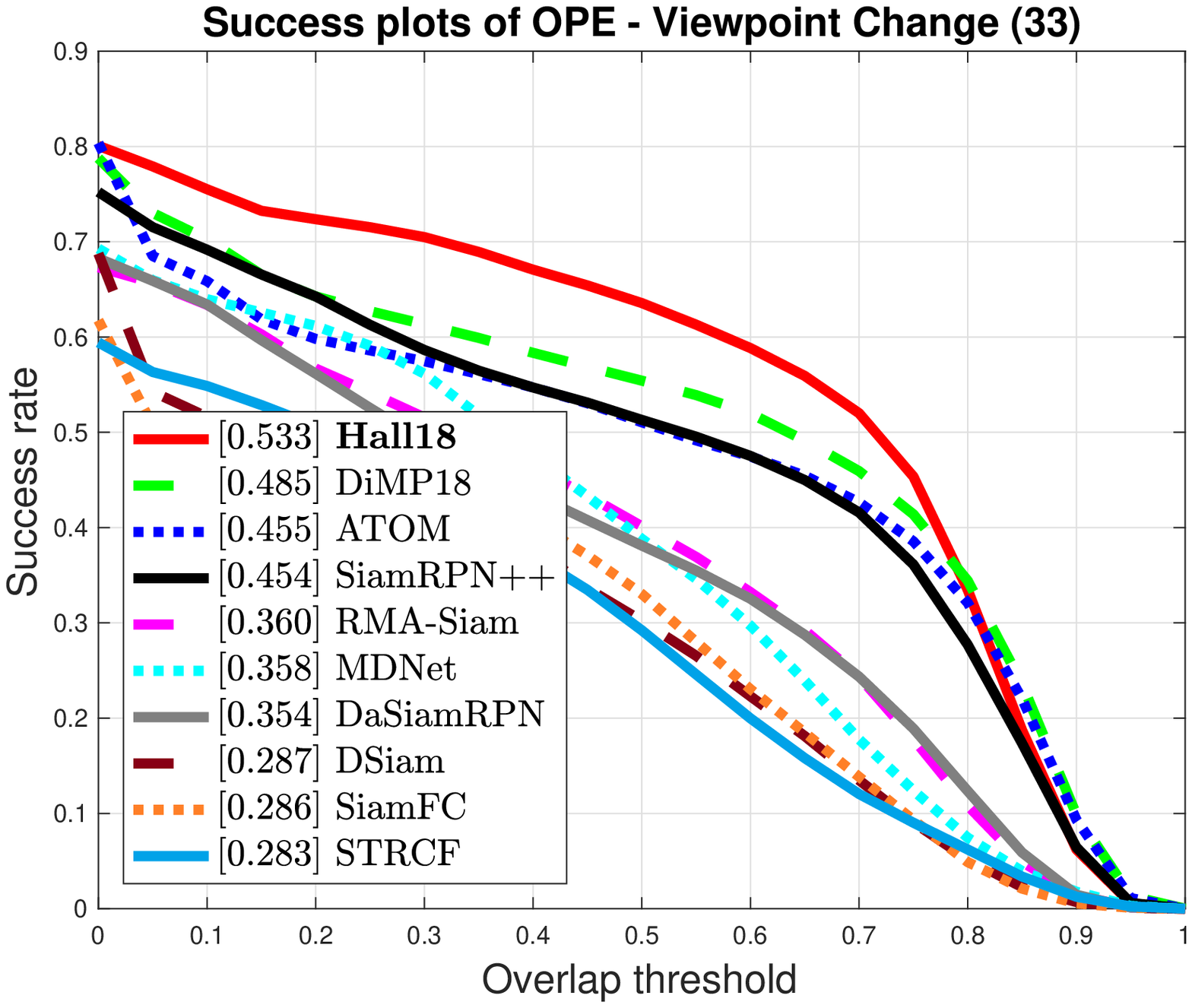}}
\subfigure{\includegraphics[width=0.23\linewidth]{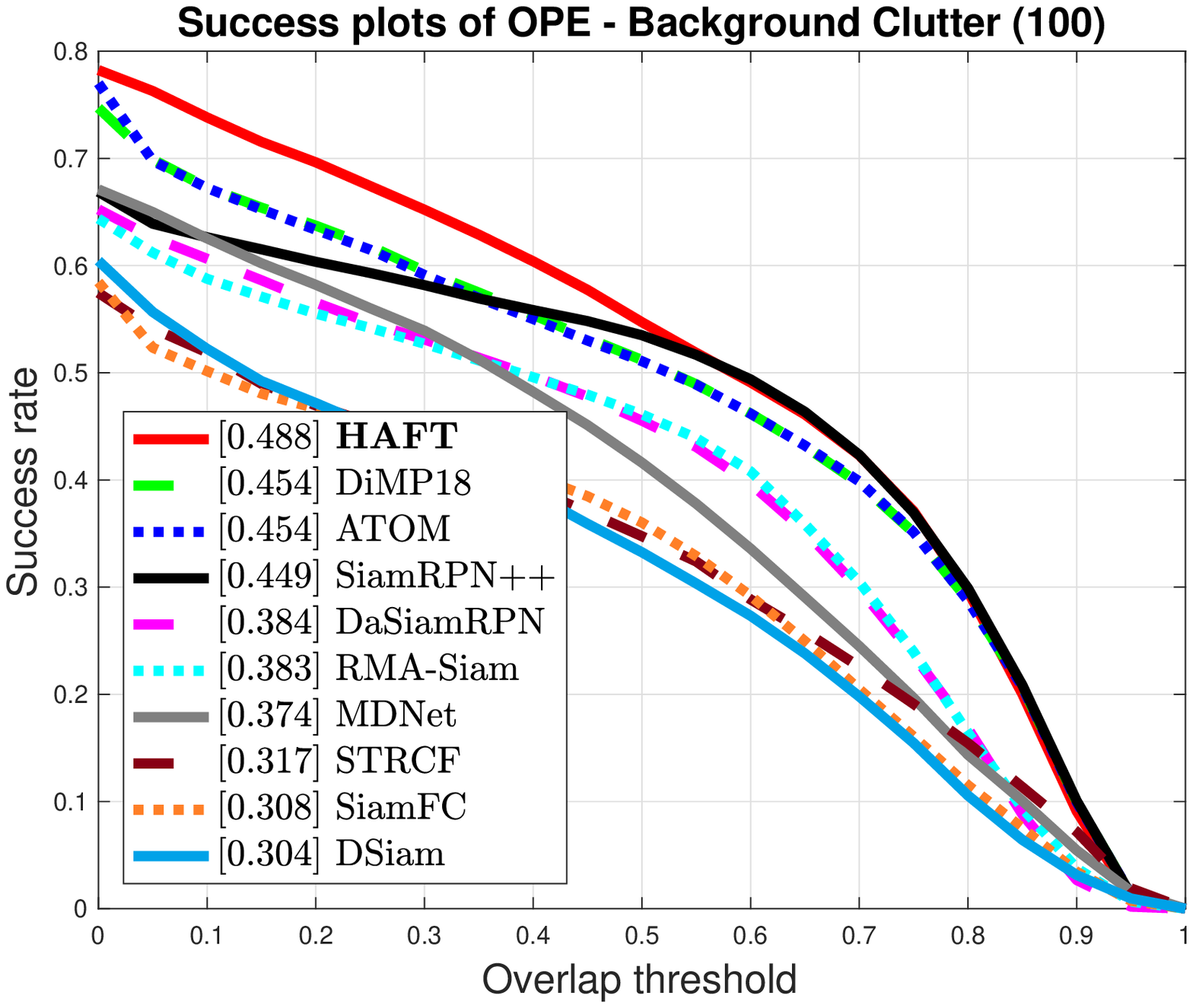}}

\end{center}
\vspace{-4mm}
\caption{The success plots of eight attributes on LaSOT dataset.}
\label{fig:lasot_attr}
\end{figure}

Further, we analyze our tracker with respect to 8 different attributes, including aspect ratio change, scale variation, partial occlusion, deformation, full occlusion, motion blur, viewpoint change, and illumination variation. As shown in Fig. \ref{fig:lasot_attr}, our tracker can handle the full occlusion problem. Compared with DiMP-18, our tracker achieves 2.2\% absolute improvement on full occlusion attributes and achieves 1.6\% absolute improvement on partial occlusion attributes. Besides, our tracker generalizes to other attributes well and achieves 2\% or so absolute improvement in aspect ratio changes, scale variation, deformation, viewpoint change, and background clutter.

\textbf{TrackingNet\cite{trackingnet}.} TrackingNet provides a large amount of data to assess trackers in the wild. We evaluate our trackers on testing dataset with 511 videos. Following\cite{trackingnet}, we use three metrices, including Precision Plots(PRE), Normalized Precision Plots(NPRE), and Success Plots(AUC), for evaluation. As show in Tab. \ref{tab:trackingnet}, our tracker achieves 0.661 on Precision Plots, 0.779 on Normalized Precision Plots, and 0.720 on Success Plots.

\setlength{\tabcolsep}{1mm}
\begin{figure}[t]\CenterFloatBoxes
\begin{floatrow}
\killfloatstyle\ttabbox[\Xhsize]
{\caption{Evaluation results of different trackers on TrackingNet. The best top 3 results are marked as \first{red}, \second{blue}, and \third{green}.}\label{tab:trackingnet}}
{
	\begin{tabular}{c | c c c c c c c | c}
	\toprule	
	Tracker & ECO 	&	SiamFC 	& MDNet & DaSiamRPN	& ATOM & SiamRPN++ & DiMP-18 & HAFT	\\
			& \cite{CVPR17ECO} & \cite{ECCV16SiamFC} & \cite{CVPR16MDNet} & \cite{eccv2018dasiamrpn}  & \cite{CVPR19ATOM} & \cite{CVPR19SiamRPN++} & \cite{iccv19dimp} & \\
	\midrule
	PRE($\uparrow$)			  &  0.492	&	0.533	 & 0.565	    & 0.591 	& 0.648   & \first{0.694}   & \second{0.666}	& \third{0.661}		\\
	NPRE($\uparrow$)	  &  0.618 	&	0.666	 & 0.705  	& 0.733		& 0.771	  & \first{0.800}	& \second{0.785}	& \third{0.779}		\\
	AUC($\uparrow$)				  &  0.554  &	0.571	 & 0.606	    & 0.638		& 0.703   & \first{0.733}	& \second{0.723}	& \third{0.720}		\\
	\bottomrule
	\end{tabular}
}
\end{floatrow}
\end{figure}

\textbf{UAV123\cite{uav}.} UAV123 dataset focus on drone low-altitude tracking. The dataset consists of 123 videos and the viewpoint is top view. We compare our method with 7 top-ranked trackers in Success Plots(AUC) and Precision Plots(PRE). As shown in Tab. \ref{tab:uav123}, our tracker is top-ranked on Success Plots, which is 0.637.

\setlength{\tabcolsep}{1mm}
\begin{figure}[t]\CenterFloatBoxes
\begin{floatrow}
\killfloatstyle\ttabbox[\Xhsize]
{\caption{Evaluation results of different trackers on UAV123. The best top 3 results are marked as \first{red}, \second{blue} and \third{green}.}\label{tab:uav123}}
{
	\begin{tabular}{c | c c c c c c c | c}
	\toprule	
	Tracker & ECO-HC 	&	SiamRPN++ 	& ECO & DaSiamRPN	& SiamRPN & ATOM & DiMP-18 & HAFT	\\
			& \cite{CVPR17ECO} & \cite{CVPR19SiamRPN++} & \cite{CVPR17ECO} & \cite{eccv2018dasiamrpn}  & \cite{CVPR18SiamRPN} & \cite{CVPR19ATOM} & \cite{iccv19dimp} & \\
	\midrule
	PRE($\uparrow$)			  &  0.725	&	0.807	 & 0.741	    & 0.796 		& 0.748   & \first{0.843}   & \second{0.836}	& \third{0.835}				\\
	AUC($\uparrow$)			  &  0.506  &	0.613	 & 0.525	    & 0.586		& 0.527   & \third{0.631}			& \second{0.632}	& \first{0.637}		\\
	\bottomrule
	\end{tabular}
}
\end{floatrow}
\end{figure}

\begin{figure}[t]
\begin{center}
\includegraphics[width=1\linewidth]{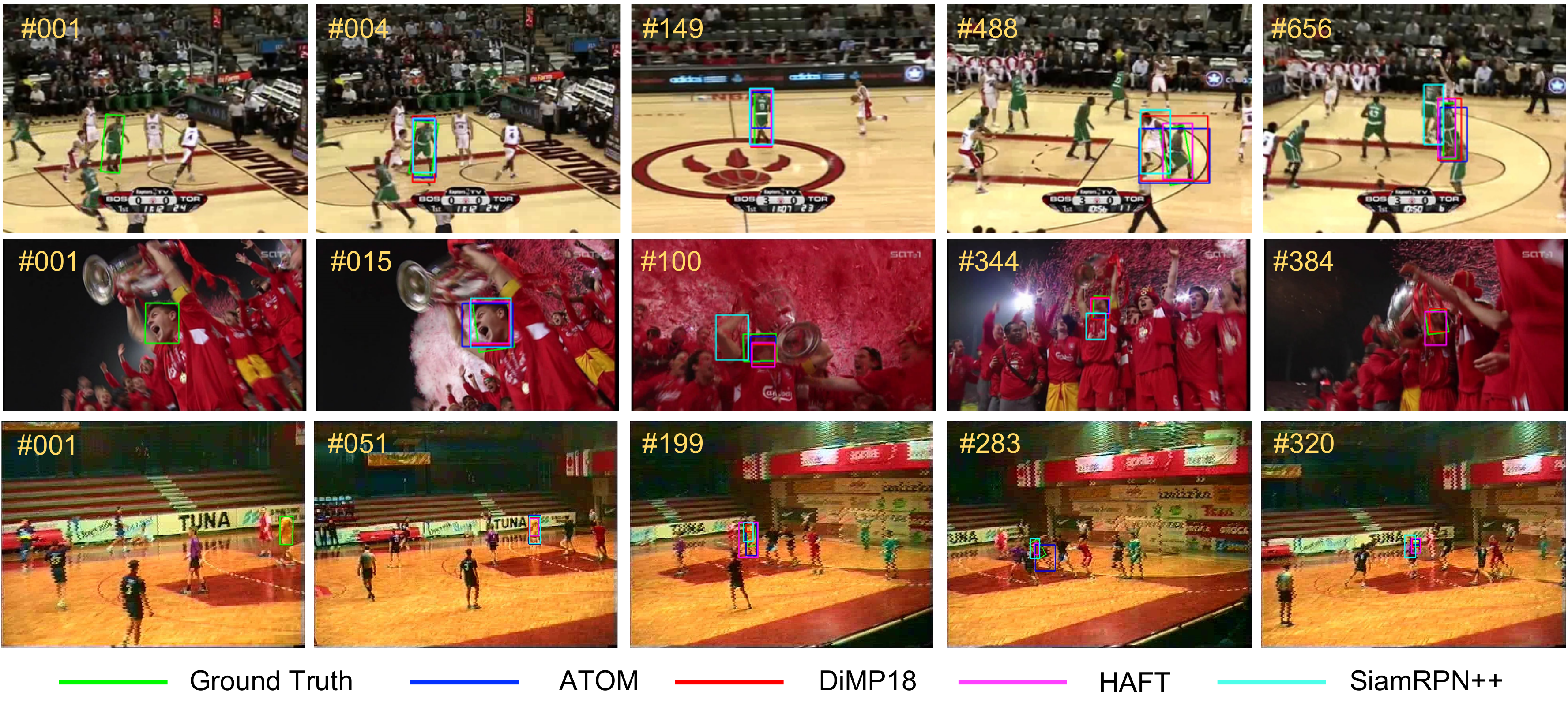}
\end{center}
\vspace{-4mm}
\caption{Visualization results. The videos in the first, second, third row are \latinword{basketball}, \latinword{soccer1}, and \latinword{handball2} in VOT2018, respectively. The bounding boxes are not drawn if the tracker lost the target (except the first frame).}
\label{fig:vis}
\end{figure}

\subsection{Qualitative Results}
Qualitative results are demonstrated in Fig. \ref{fig:vis}. The first row in the figure is the first frame of the video. The tracking target is surrounded by the green bounding box. In the subsequent frames, the pink bounding box is our tracking algorithm. Compared with DiMP-18, our tracker can better handle the full occlusion problem. Especially, the \latinword{soccer1} video, the lost number of HAFT is less than the compared tracker. 

\section{Related Work}

The past years have seen large improvements in visual object tracking, thanks to powerful baselines such as correlation filter based algorithm\cite{CVPR17ECO,PAMI15KCF}, Siamese network based algorithms\cite{ECCV16SiamFC,CVPR18SiamRPN}, and the unified online learning and offline training approaches\cite{CVPR19ATOM,iccv19dimp}.

The CF based methods dedicate to fast online learning and incorporate different features.Henriques \etal\cite{PAMI15KCF} introduced circular matrix to speed up the online learning of CF methods. Bertinetto \etal\cite{CVPR16Staple} extended with color histogram to compensate the HOG feature which loses the color information. Danelljan \etal\cite{ECCV16CCOT} converted different kinds of features to continuous domain to tackle combining inconsistent resolutions of features. The drawback of CF based method is that we cannot learning features that suit for object tracking. Siamese approaches start drawing much attention as this time. Bertinetto \etal\cite{ECCV16SiamFC} brought cross correlation into a fully convolutional network which learned the tracking feature representation. Li \etal\cite{CVPR18SiamRPN} borrowed the region proposal network\cite{ren2015faster} to Siamese network, further increasing tracking accuracy and tracking speed. Li \etal\cite{CVPR19SiamRPN++} made the Siamese network going deeper with the spatial-aware sampling strategy. Danelljan \etal\cite{iccv19dimp} unified the offline training tracking representation with the Siamese network and online learning with steepest gradient decent.

A key limitation in the above method is that they lack of motion information which easily fails in occlusion situation. Gladh \etal\cite{icpr16deepmotion} first proposed fusing appearance cues with deep motion cues with optical flow. The limits is that extracting optical flow which further send to a deep action recognition network is time consuming. Zhu \etal\cite{cvpr18flowtrack} motivated by DFF\cite{zhu2017deep} use the FlowNet\cite{cvpr15flownet} to propagates deep features of previous frames to subsequent frames. However, when the object is occluded, the extracted flow field is inaccurate. Yang \etal\cite{wang2019prediction} incorporated Kalman Filter to estimate object position which is simple but effective. The drawback is that it cannot utilize large video dataset to learn a dynamic temporal model and limits its further improvement.

In our work, we utilize a conditional GAN\cite{cgan} for deep future representation generation. A few number of GAN approaches can be found for visual object tracking\cite{cvpr18vital,cvpr2018sint++}. Wang \etal\cite{cvpr2018sint++} directly used the generative network to sample massive hard positive samples with deep reinforcement learning to decide the mask position. Song \etal\cite{cvpr18vital} augmented positive samples with generative network to randomly choose predefined masks. Instead of using the GAN to generative more hard positive samples, our approach utilize the GAN to generate future frame representation.

\section{Conclusion}
In this paper, our model mimics the human biological process. Our future frame anticipation method learns to anticipate future scene representations while predicting the future movement of the target. Generative adversarial loss with $L_2$ loss make the forecasted future frame embedding close to the realistic ones.  The resulting tracker benefits from the robust hallucinate features. Our qualitative quantitative results demonstrate the superior performance even under severe occlusion. It achieves promising results on VOT2018, LaSOT, OTB100, TrackingNet, and UAV123.

\clearpage

\bibliographystyle{nips}
\bibliography{egbib}
\end{document}